\documentclass[11pt]{article}
\usepackage{amsmath}
\usepackage{amssymb}

\newcommand{\mathbold}[1]{\mbox{\boldmath $\bf#1$}}

\renewcommand{\glossary}[2]{}
\def\addcontentsline#1#2#3{} 
\newtheorem{thm}{Theorem}

\newcommand{\E}{\mathbold{E}}

\newcommand{\be}{\begin{equation*}}

\renewcommand{\hbar}{{\overline h}}

\newcommand{\OTS}{{\text{\rm OTS}}}

\newcommand{\pibar}{{\bar \pi}}

\begin{document}
\title{Some observations concerning Off Training Set (OTS) error.}
\author{Jonathan Baxter \\ 
        Department of Systems Engineering \\ 
	Australian National University \\ 
	Canberra 0200, Australia}
\date{August 16, 1999}
\maketitle

\abstract{A form of generalisation error known as 
{\em Off Training Set} (OTS) error was recently introduced
in \cite{wolly96a}, along with a theorem showing that small training 
set error does not guarantee small OTS error, unless assumptions 
are made about the target function. Here it is shown that the applicability 
of this theorem is limited to models in which the distribution 
generating training data has no overlap with the distribution generating 
test data. It is argued that such a scenario is of limited relevance to 
machine learning.} 

\section{Introduction}
A new measure
of generalisation error called Off Training Set (OTS) error
was introduced recently in \cite{wolly96a,wolly96b}. Under
quite weak assumptions it was shown that with respect to OTS error there
are no {\em a priori} distinctions between learning algorithms, at least
if it is assumed that the target functions are uniformly distributed.
Thus, as far as OTS error is concerned, an algorithm that minimizes 
error on the training set will do no better than random guessing. 
If OTS error accurately models the concept of generalization then this 
is a depressing conclusion indeed.

However, in this paper it is argued that OTS error 
does not model what is normally
meant by generalization error.
In particular, it is shown that the assumptions underlying
one of the main ``no free lunch'' (NFL) theorems (theorem 2) in
 \cite{wolly96a} imply that the distributions
used to generate training data and testing data have
disjoint supports. Thus, training a neural network to recognise faces by 
showing it images of handwrittten characters 
is the kind of learning problem covered by 
the NFL theorem. Not surprisingly, one cannot conclude anything about 
generalisation performance in such circumstances, but it would seem that 
such a scenario is of little interest in machine learning and statistics 
anyway.

\section{OTS error}
\label{OTS}
For simplicity of exposition, the following restricted learning
scenario is considered. In the notation of \cite{wolly96a}, the
learning algorithm is supplied with a training set $d =
\{d_X(i),d_Y(i)\}$, $i=1,\dots,m$, where each $d_X(i)$ is
chosen from the (finite) input space $X$ according to a distribution
$\pi$, and $d_Y(i) = f(d_X(i))$ 
is some fixed Boolean target function, $f\colon X\to
\{0,1\}$.  The set of all $d_X(i) \in d$ is denoted by $d_X$.  
The learning algorithm is assumed to be deterministic,
so that in response to the taining set $d$,
the algorithm produces a hypothesis
$h_d\colon X\to \{0,1\}$. 

The generalization performance of the algorithm is 
measured by the {\em off training set error} (OTS error):
$$
E_\OTS(d,f) := \frac{1}{\sum_{x\in X-d_X} \pi(x)}
\sum_{x\in X - d_X} \pi(x) |h_d(x) - f(x)|.
$$
Note that OTS error is just the expected error of the algorithm's
hypothesis on those inputs {\em not} appearing in the training set. 
Another way of expressing OTS error is as the expected loss of the learner 
with respect to the {\em testing distribution}: $\pibar_d(x) := 0$ if 
$x \in d_X$ and $\pibar_d(x) := \pi(x)/\sum_{x\in X-d_X} \pi(x)$ 
if $x\notin d_X$.
Note that $\pibar_d$ depends on the training set $d$. 
The more general case where 
$\pibar_d(x)$ is any distribution on $X$ with the property that 
$\pibar_d(x) = 0$ if $x\in d_X$ is also considered in \cite{wolly96a} 
(see the remarks preceding theorem 2 in that paper). In either case we 
can write,
\begin{equation}
\label{ots}
E_\OTS(d,f) = \sum_{x\in X} \pibar_d(x) |h_d(x) - f(x)|.
\end{equation}
As the input space is finite and we are only considering Boolean target 
functions, there is no difficulty with the concept of choosing a target 
function $f$ uniformly at random. The uniform average over all target 
functions will be denoted by $\E_f$. The following theorem is essentially
theorem 2 from \cite{wolly96a}, applied to the particular scenario of the 
present paper. 

\begin{thm}
\label{bong}
Suppose that $P(d|f)$ is independent of $f(x)$ for all  $x\in X - d_X$
(such a $P(d|f)$ is called a {\em vertical likelihood} in \cite{wolly96a}). 
Furthermore, suppose that for all training sets $d$ of size $m$,
the testing distribution $\pibar_d(x) = 0$ if $x\in d_X$. Then,
\begin{equation}
\label{bongeq}
\E_d \E_f E_\OTS(d,f) = \frac12,
\end{equation}
where $\E_d$ is the expectation over all training sets $d$ of size $m$.
\end{thm}

\section{Discussion}
On face value, theorem \ref{bong} looks rather negative. It says that
{\em any} algorithm for choosing a hypothesis $h_d$ based on 
training data $d$ will have an expected OTS error of 1/2. As random guessing 
would give an expected error of 1/2, theorem \ref{bong} would appear to show
that no algorithm can do better than random guessing. 

However, closer inspection reveals that while OTS error is a reasonable
candidate for generalisation error when considered in the context of 
a {\em single} training set $d$, it is quite pathological when expectations
are taken over {\em all} training sets, as is the case in theorem \ref{bong}.
Specifically, substituting \eqref{ots} into the left hand side of
\eqref{bongeq} gives
\begin{equation}
\label{boing}
\E_d \E_f \sum_{x\in X} \pibar_d(x) |h_d(x) - f(x)| = \frac12.
\end{equation}
As the testing distribution $\pibar_d$ varies with $d$,
this expression cannot be interpreted as the expected OTS error of the
algorithm with respect to some {\em fixed} testing distribution. A test 
distribution that depends on the training data is too hard a target 
for machine learning because it encompasses the situation in which
an adversary generates training sets according to some fixed
distribution $\pi$, but then varies the distribution generating test sets 
depending on the particular training set produced.

For theorem \ref{bong} to be more relevant to machine learning
a {\em fixed} test distribution $\pibar(x)$ must be chosen.  
However, one of the
crucial assumptions in the proof of theorem \ref{bong} is that the
testing distribution satisfies $\pibar(x) = 0$ if $x\in d_X$, for {\em
any} training set $d$ of size $m$ of positive probability (see
\cite{wolly96a}, appendix C. The condition used there is actually that
$\pibar(x) = 0$ for any training set $d$ of size $m$ (regardless of whether 
it has positive probability), but the theorem
still holds under the weaker assumption above). This implies that
$\pibar(x) = 0$ for any $x$ such that $\pi(x) >0$ (recall that
$\pi(x)$ is the probability of input $x$ appearing in the training
set). In other words, for a fixed test distribution $\pibar$, the assumptions 
behind theorem
\ref{bong} imply that the training distribution $\pi$ and the
testing distribution $\pibar$ have disjoint supports. This means that 
no matter how large the training set is, there is always {\em zero}
probability of seeing an example in testing that was already seen in 
training. 

Clearly under such circumstances one cannot conclude anything about the 
generalisation behaviour of a learning algorithm, which is the content of 
theorem \ref{bong}. However, disjoint training and testing distributions is
unlikely to be interesting from a machine learning perspective. 
Some kind of relationship between training and test data is always 
assumed, otherwise there would be no point
feeding the training data into the learning algorithm in the first place.
In fact in practice, where possible, 
the assumption that the training and testing data 
are generated according to the same distribution is usually {\em 
engineered}.

Put another way, no-one would train a neural network to
recognize faces by feeding it a training set consisting of
images of handwritten characters.

\subsection*{Very large input spaces}
If the input space $X$ is very large then in practice the training set and
testing set will almost always be disjoint, even if the training and testing 
distributions are identical. Under such 
circumstances one might expect the negative conclusion of theorem \ref{bong}
to apply. However, it does not, the reason being the subtle 
difference between ``almost never seeing the same example in testing 
as seen in training'' and 
ruling out a-priori any possibility of seeing the same example in testing
as seen in training. The latter has to hold if the NFL theorems are to apply.

\section{Conclusion} 
We have seen that the negative conclusions of the ``No Free Lunch'' theorems
can be avoided by assuming that the training and test distributions have
some overlap.  Note that although we have to assume something about the 
input-space distributions,  we do not have to assume anything about the 
distribution over target functions.  

\bibliographystyle{apalike}
\bibliography{bib}

\end{document}